\documentclass[letterpaper, 10 pt, conference]{ieeeconf}  % Comment this line out if you need a4paper [letterpaper]

\IEEEoverridecommandlockouts                              % This command is only needed if 
\usepackage{graphicx}
\usepackage{tabularx}
\usepackage{subcaption}
\usepackage{rotating}
\usepackage{colortbl}
\usepackage{array}
\usepackage{pifont}
\usepackage{arydshln}
\usepackage[font=small,labelfont=bf]{caption}
\usepackage{mwe}
\usepackage{amsmath}
\usepackage{tablefootnote}
\usepackage{url}
\usepackage{listings}
\usepackage{booktabs}
\usepackage{siunitx}
\usepackage{float}
\usepackage{multirow}
\usepackage{hhline} % For double lines
\usepackage{tablefootnote}
\usepackage{dingbat}
\usepackage{tikz}
\usepackage{pgf}
\usepackage{bbding}
\usepackage{color}
\usepackage{cuted}
\usepackage{amssymb}
\usepackage{capt-of}
\usepackage{makecell}
\definecolor{cvprblue}{rgb}{0.21,0.49,0.74}
\usepackage{booktabs}
\newcommand{\rot}[1]{\rotatebox{90}{#1}}
\newcommand{\greencheck}{\textcolor{green!70!black}{\ding{51}}}
\newcommand{\redcross}{\textcolor{red}{\ding{55}}}
\usepackage[pagebackref,breaklinks,colorlinks,citecolor=cvprblue]
{hyperref}
\usepackage{array}
\title{\LARGE \bf
 Pedestrian Intention and Trajectory Prediction in Unstructured Traffic Using IDD-PeD
}

\author{Ruthvik Bokkasam$^{1}$, Shankar Gangisetty$^{1}$, A. H. Abdul Hafez$^{2}$, and C. V. Jawahar$^{1}$% <-this % stops a space
\thanks{$^{1}$IIIT, Hyderabad, India
{\tt\small ruthvik.cvresearch@gmail.com, \{shankar.gangisetty@ihub-data.,  jawahar@\}iiit.ac.in}}%
\thanks{$^{2}$King Faisal University, Al Hofuf,
Saudi Arabia
{\tt\small aabdulhafiz@kfu.edu.sa}}%
}
\begin{document}
\maketitle

% \begin{strip}
% \centering
% %\vspace{-.95cm}
% \includegraphics[width=1\linewidth]{images/Pedestrian_Intent_Prediction.png}
% \captionof{figure}{caption for teaser fig~\ref{tab:comparison_table}. 
% \label{fig:teaser_fig}}
% \end{strip}

\begin{strip}
    \centering
    \includegraphics[width=\textwidth]{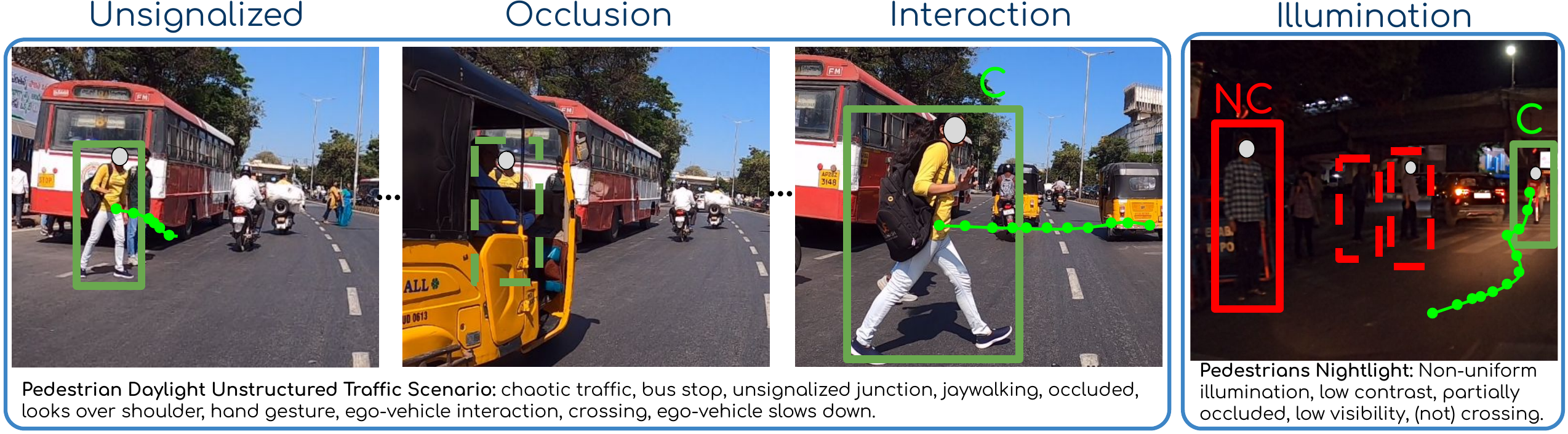} 
    \captionof{figure}{Illustration of pedestrian intention and trajectory encountering various challenges within our unstructured traffic IDD-PeD dataset. The challenges include occlusions, signalized types, vehicle-pedestrian interactions, and illumination changes. A comparative
analysis of these challenges with other datasets is presented in Table~\ref{tab:dataset_comparison}. Intent of \textcolor{green}{C:Crossing} with trajectory and \textcolor{red}{NC:Not Crossing}.}
    \label{fig:teaser}
\end{strip}
\vspace{-.5cm}
\begin{abstract}

With the rapid advancements in autonomous driving, accurately predicting pedestrian behavior has become essential for ensuring safety in complex and unpredictable traffic conditions. The growing interest in this challenge highlights the need for comprehensive datasets that capture unstructured environments, enabling the development of more robust prediction models to enhance pedestrian safety and vehicle navigation.
In this paper, we introduce an Indian driving pedestrian dataset designed to address the complexities of modeling pedestrian behavior in unstructured environments, such as illumination changes, occlusion of pedestrians, unsignalized scene types and vehicle-pedestrian interactions. The dataset provides high-level and detailed low-level comprehensive annotations focused on pedestrians requiring the ego-vehicle's attention.
Evaluation of the state-of-the-art intention prediction methods on our dataset shows a significant performance  drop of up to $\mathbf{15\%}$, while trajectory prediction methods underperform with an increase of up to $\mathbf{1208}$ MSE, defeating standard pedestrian datasets. Additionally, we present exhaustive quantitative and qualitative analysis of intention and trajectory baselines. We believe that our dataset will open new challenges for the pedestrian behavior research community to build robust models. Project Page: \url{https://cvit.iiit.ac.in/research/projects/cvit-projects/iddped}

\end{abstract}

%%%%%%%%%%% 
%%%%%%%%% BODY TEXT

\section{Introduction}\label{sec:intro}
%Autonomous vehicles, or ADAS systems, must interact smoothly with pedestrians—avoiding collisions, allowing safe crossings, and promoting efficient movement, especially in dense and dynamic environments. 
%
%%%%%%%%%% Parag 1
Autonomous vehicles and ADAS must smoothly interact with pedestrians, avoiding collisions, ensuring safe crossings, and promoting efficient movement, especially in dense and dynamic environments~\cite{Katyal2020Intent,Liu2023Intentionaware}.
%
%%%%In structured traffic settings, this can be achieved more easily through established infrastructure like traffic signals and crosswalks. However, in unstructured traffic environments, where such infrastructure is absent, pedestrian behavior becomes highly unpredictable. Here, ego-vehicles must not only predict pedestrian movements in real-time but also anticipate sudden changes in direction, speed, or intent. 
%
%In structured traffic settings, 
In traffic environments, well-established infrastructure like signals and crosswalks simplify pedestrian behavior prediction. However, the lack of such infrastructure makes pedestrian behavior far more unpredictable~\cite{Parikh2024IDDX}. Ego-vehicles must be capable of anticipating sudden changes in pedestrian direction, speed, or intent as real-time predictions of pedestrian intentions and trajectories are critical for enhancing road safety and ensuring smooth navigation in these chaotic and unpredictable traffic settings.
%The presence of well-established infrastructure in traffic environments like signals and crosswalks makes this easier. However, the absence of such infrastructure %unstructured environments makes pedestrian behavior unpredictable~\cite{Parikh2024IDDX}. Also ego vehicles must anticipate sudden changes in pedestrian direction, speed, or intent.
%
%These real-time predictions of pedestrian behaviours, including their intentions and trajectories, are essential for enhancing road safety and ensuring smooth navigation in the often chaotic and unpredictable nature of unstructured traffic conditions.
%
%Real-time predictions of pedestrian behaviors, including intentions and trajectories, are crucial for improving road safety and ensuring smooth navigation in chaotic and unpredictable unstructured traffic environments.

%%%%%%% Paragraph2
%Recent advances in deep learning have led to significant improvements in performance for perceiving and predicting the motion of dynamic objects in structured or constrained settings. 
Recent advances in deep learning have significantly improved the perception and prediction of dynamic object motion in structured settings, with several datasets developed for predicting pedestrian intentions and trajectories. Datasets like Argoverse~\cite{wilson2021aroverse2}, nuScenes~\cite{caesar2020nuscenes}, IDD-3D~\cite{Dokania2023IDD3d} and Waymo~\cite{sun2020scalability} focus on dynamic agent interactions but are primarily designed for driving perception and planning, not pedestrian intention and navigation. Datasets such as UCY~\cite{lerner2007crowds}, ETH~\cite{pellegrini2009eth}, and SSD~\cite{robicquet2016learning} provide pedestrian trajectories from top-down views, lacking driver-pedestrian interaction insights on the ground. Although vehicle-pedestrian interaction datasets like JAAD~\cite{rasouli2017they}, PIE~\cite{rasouli2019pie}, TITAN~\cite{malla2020titan},  PSI~\cite{chen2021psi}, Euro-PVI~\cite{bhattacharyya2021euro} and TBD~\cite{Wang2024TBD} target intention and trajectory prediction, they are tailored to structured environments with signals and crosswalks, and do not account for diverse pedestrian behaviors in unstructured settings.

On chaotic urban streets without signals, autonomous vehicles face pedestrian behaviors like jaywalking or sudden halts. Anticipating these actions is crucial for safe navigation and collision avoidance, as conventional traffic rules often don't apply in such settings. Considering the complexity of pedestrian behavior in unstructured traffic environments, we identify four key factors that impact prediction accuracy: (i) \textit{occlusions}, (ii) \textit{lighting conditions}, (iii) \textit{signalized type}, and (iv) \textit{vehicle-pedestrian interactions}. These factors add unpredictability, demanding robust models to handle such diverse scenarios. However, most existing datasets overlook these nuances, limiting their effectiveness in modeling pedestrian behavior in these challenging conditions.

\begin{figure}%[h]
    \centering
    \includegraphics[width=\columnwidth]{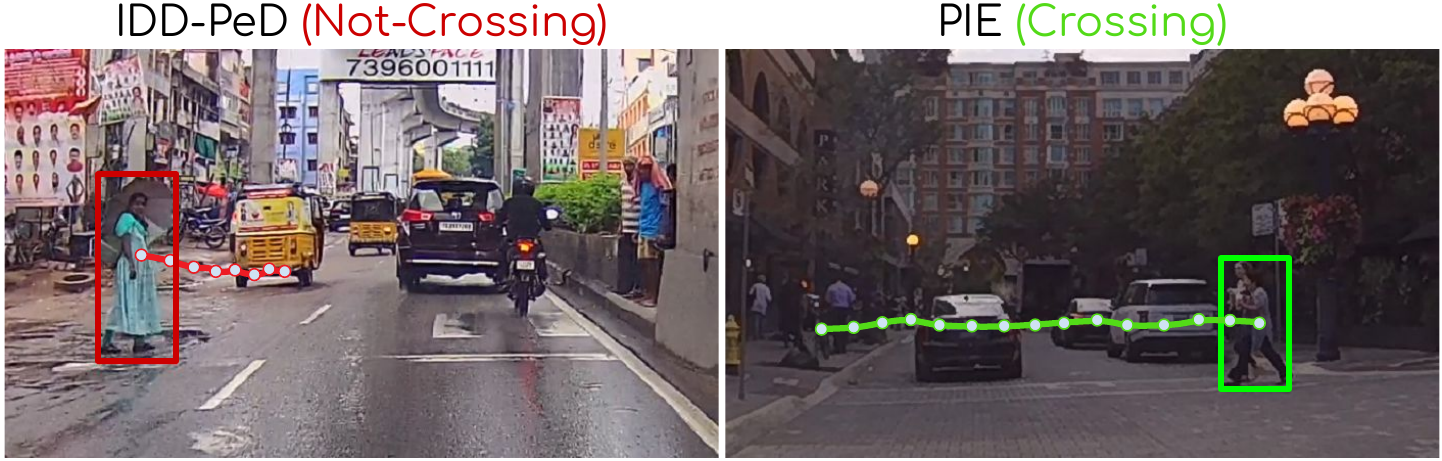} 
    \caption{\textbf{Rolling behavior} is common in unstructured environments, where pedestrians hesitate, pause, or change direction unpredictably, even after they start to cross. \textit{(left)} In our dataset, due to rolling behaviour, the pedestrian does \textcolor{red}{not cross} and the ego-vehicle moves ahead, while \textit{(right)} in PIE, the pedestrian \textcolor{green}{crosses} since structured settings typically prompt the ego-vehicle to yield giving priority to the pedestrian.
    %Rolling behavior is common in unstructured traffic environments, where pedestrians hesitate, pause, or change direction unpredictably. (Left) In our dataset, PIP predicts "not crossing" due to this behavior with vehicle making its path, while (right) in PIE, "crossing" is predicted since structured settings typically prompt the ego-vehicle to yield giving priority to pedestrian.
    %Future vehicle locations are displayed in \protect\bluecirclesymbol.
    %where a pedestrian may start crossing the road but doesn't end up crossing in front of the ego-vehicle. This corresponds to a negative label for pedestrian intention prediction. Structured datasets like PIE have no such instances since such environments are characterized by vehicles yielding to pedestrians giving them higher priority. Blue indicates the future path of the vehicle.
    }
    \label{fig:Ours_PIE}
    \vspace{-.5cm}
\end{figure}

%Building on this, 
Now, let us understand these four key factors that are crucial within unstructured traffic conditions, see scenarios in Fig.~\ref{fig:teaser} and comparison in Table~\ref{tab:dataset_comparison}): 
(i) \textit{Occlusion}: Dense and dynamic traffic often occludes pedestrians, and datasets like TITAN~\cite{malla2020titan} attempt to bypass this by removing occluded regions, which degrades pedestrian intent and trajectory predictions. 
(ii) \textit{Illumination variation}: The variability in lighting conditions can significantly impact pedestrian behaviour predictions. PIE and TITAN lack nightlight data, while a dataset like NightOwls~\cite{neumann2019nightowls} lacks daylight data, degrading the performance across varying lighting conditions. 
(iii) \textit{Signalized type}: The absence of traffic signals and crosswalks especially in unstructured settings lead to unpredictable pedestrian-vehicle interactions. JAAD and PIE are captured in scenes with well-established infrastructure including signals and crosswalks simplifying pedestrian behaviour predictions. 
(iv) \textit{Vehicle-Pedestrian Interactions}: Datasets like PIE and JAAD lack interaction information, which reduces their effectiveness in modeling pedestrian behaviors. In contrast, unstructured environments present more frequent and complex interactions~\cite{Mukoya2024Jaywalker}. 
%For instance, “rolling behavior”~\cite{vasudevan2020pedestrian} involves pedestrians continuously adjusting their speed and path in response to traffic, increasing interaction frequency. As seen in Fig.~\ref{fig:Ours_PIE}, poor traffic rule enforcement in unstructured environments results in low vehicle yielding, even near crosswalks. Although datasets like TITAN~\cite{malla2020titan} and Euro-PVI~\cite{bhattacharyya2021euro} include some instances of vehicle-pedestrian interaction, they lack detailed annotations, limiting their effectiveness in modeling these behaviors.
%
For example, “rolling behavior”~\cite{vasudevan2020pedestrian} describes pedestrians constantly adjusting their speed and path to navigate traffic, leading to more frequent interactions. As shown in Fig.~\ref{fig:Ours_PIE}, weak traffic rule enforcement in unstructured environments results in minimal vehicle yielding, even at crosswalks. While TITAN~\cite{malla2020titan} and Euro-PVI~\cite{bhattacharyya2021euro} capture vehicle-pedestrian interactions, they lack explicit annotations, limiting their ability to model these behaviors accurately.

To bridge these gaps, we introduce the IDD-PeD dataset. The dataset contains over $650K$ annotated bounding boxes and provides comprehensive annotations for pedestrian intention, location, ego-vehicle motion, scene context, occlusion, and pedestrian-vehicle interactions. By capturing these factors, IDD-PeD supports the development of robust models capable of navigating complex, real-world unstructured environments. 
Fig.~\ref{fig:teaser} depicts samples from the proposed dataset comprising of the diverse challenges and Table~\ref{tab:dataset_comparison} compares our dataset over the existing datasets addressing these challenges.
%As shown in Table~\ref{tab:dataset_comparison}, our dataset addresses key limitations in previous benchmarks, offering diverse lighting conditions, detailed vehicle-pedestrian interaction data, and full occlusion annotations.
In summary, we make the following contributions:
\begin{itemize}
    \item We introduce the largest dataset, i.e., IDD-PeD, with more than $650K$ pedestrian bounding boxes and $19$ behavioral attributes, captured in dense and unstructured traffic environments. The dataset addresses key challenges such as \textit{Occlusions}, \textit{Lighting Conditions}, \textit{Signalized Types}, and \textit{Vehicle-Pedestrian Interactions} (see Fig.~\ref{fig:teaser} and Table~\ref{tab:dataset_comparison}).
    % \item \textcolor{blue}{mentioning contribution on annotations will be good (if any)}
    \item We benchmark eight baseline methods for pedestrian intention prediction (PIP) and four trajectory prediction (PTP) against the IDD-PeD dataset, showing a performance drop of up to $15\%$ in PIP (see Table~\ref{tab:pip_models}) and an increase in MSE of up to $1208$ for deterministic models and $224$ for stochastic models in PTP (see Table~\ref{tab:ptp_models}) compared to previous benchmark datasets.
    %We benchmark the IDD-PeD dataset against eight pedestrian intention prediction (PIP) baseline methods and four trajectory prediction (PTP) baseline methods,
    %, including PIETraj~\cite{rasouli2019pie} and SGNet~\cite{wang2022stepwise} for pedestrian trajectory prediction, 
    %showing a performance drop of up to \_ for PIP and \_ for PTP compared to standard benchmark datasets.
   % \item We provide comprehensive quantitative analysis of key factors such as \textit{lighting conditions, occlusion, signalized type, and vehicle-pedestrian interactions}, revealing their significant impact on model robustness (see Tables IV and V).
    \item We present a comprehensive quantitative and qualitative analysis of the key challenges %, including \textit{lighting conditions, occlusion, signalized type, and vehicle-pedestrian interactions}, 
    highlighting their substantial impact on model robustness (see Tables~\ref{tab:pip_comparisons} and~\ref{tab:ptp_comparisons}).%, as well as qualitative analysis for the same (see Fig.~\ref{fig:ptp_qualitative}).
\end{itemize}

\begin{table}%[h]
\centering
\footnotesize
\setlength{\tabcolsep}{1.4pt}  % Increased space between columns
\def\arraystretch{1.2}
\newcolumntype{R}{>{\raggedleft\arraybackslash}p{1.4cm}}
\begin{tabular}{R *{13}{c}}
\toprule
\multirow{2}{*}{{Dataset}} & \rot{\#Pedestrians} & \rot{\makecell{\#Bounding \\Boxes}} & \rot{\makecell{\#Annotated \\ Frames}} & \rot{\makecell{\#Behavioral \\Classes}} & \rot{\makecell{Traffic \\Density}} & \rot{\makecell{OBD}} & 
\rot{\makecell{Video Seq.}} & \rot{\makecell{Group}} & \rot{\makecell{Location}} & \rot{\makecell{Occlusion}} & \rot{\makecell{Nightlight}} &
\rot{{Signalized}} &
\rot{{Interaction}} 
%& \rot{\makecell{Rolling \\Behavior}} 
\\
% & \rot{\textbf{Pedestrians}} & \rot{\textbf{Bboxes}} & \rot{\textbf{Frames}} & \rot{\textbf{classes}} & \rot{\textbf{Density}} & \rot{\textbf{Data}} & \rot{\textbf{Annotations}} & \rot{\textbf{Annotations}} & \rot{\textbf{Annotations}} & \rot{\textbf{Annotations}} & \rot{\textbf{Annotations}} & \rot{\textbf{Video seq.}} & \\
\midrule
JAAD~\cite{rasouli2017they} & 2.8K & 391K & 75K & 11 & Low & \redcross & \redcross & \greencheck & \redcross & \greencheck & \greencheck & \greencheck & \redcross \\ %& \redcross 

PIE~\cite{rasouli2019pie} & 1.8K & 750K & 293K & 11 & Low & \greencheck & \greencheck & \redcross & \redcross & \greencheck & \redcross & \greencheck & \redcross \\ %& \redcross 

TITAN~\cite{malla2020titan} & 8.6K & NA & 75K & 33 & High & \greencheck &  \redcross& \redcross & \redcross & \redcross & \redcross & \redcross & \redcross \\ %& \redcross \\

STIP~\cite{liu2020spatiotemporal} & 3.3K & 350K & 110K & -- & Medium & \redcross  & \greencheck & \redcross & \redcross & \redcross & \redcross* & \redcross &  \redcross \\ %& \redcross \\

PSI~\cite{chen2021psi} & 180 & NA & 26K & -- & Low & \redcross & \redcross & \redcross & \redcross & \greencheck & \redcross* & \redcross & \redcross \\ %& \redcross \\

\hdashline
\textbf{IDD-PeD (Ours)} & 5K & 686K & 205K & 19 & High & \greencheck & \greencheck & \greencheck & \greencheck & \greencheck & \greencheck & \greencheck & \greencheck \\ % & \greencheck \\

\bottomrule
\end{tabular}
\\ * day and night not explicitly mentioned
\caption{\textbf{Comparison of datasets for pedestrian behavior understanding.}
%Interaction refers to a binary indicator of influence between the ego-vehicle and pedestrian.
On-board diagnostics (OBD) provides ego-vehicle speed, acceleration and GPS information. Group annotation represents the number of pedestrians moving together, in our dataset, about $1,800$ move individually while the rest move in groups of $2$ or more. Interaction annotation refers to a label between ego-vehicle and pedestrian, where both influence each other's movements and decisions. \greencheck and \redcross ~indicate presence or absence of annotated data.}
\label{tab:dataset_comparison}
\vspace{-.5cm}
\end{table}

%%%%%%%%%%%%%%%%%%%%%%%%%%%%%%%%%%%%
\section{Related Work}
%In this section, we discuss two types of pedestrian behavior prediction techniques: PIP and PTP methods. %pedestrian intention prediction and trajectory prediction methods. 
%We then analyze the datasets used to evaluate these methods.

%%%%%%%%%%%%%%%%%%%%%%%%%%%%%%%%%%%%%%%%%%%%%%%%%%%%%%
\subsection{Pedestrian Intention Prediction}
Early works in PIP have evolved from single-frame CNN predictions \cite{rasouli2017they}\cite{simonyan2014very} into two main categories: spatio-temporal modeling and feature fusion. Spatio-temporal models use CNNs to extract visual features, followed by RNNs for temporal reasoning \cite{liu2020spatiotemporal}\cite{lorenzo2020rnn}\cite{kotseruba2020they}, with methods like PIE \cite{rasouli2019pie} using ConvLSTMs. Other approaches leverage 3D CNNs, such as 3D DenseNet \cite{saleh2019real}\cite{saleh2020spatio}, C3D \cite{tran2015learning}, and I3D \cite{carreira2017quo}, or incorporate pose estimation with 2D pose features or Graph Convolutional Networks (GCNs) \cite{sucha2017pedestrian}\cite{cadena2019pedestrian}.

%Recent works in PIP have evolved from predictions from a single frame using CNNs \cite{rasouli2017they}\cite{simonyan2014very} into two main categories: spatio-temporal modeling and feature fusion methods. Several studies have employed spatio-temporal modeling by using CNNs to extract visual features, followed by RNNs for temporal reasoning \cite{liu2020spatiotemporal}\cite{lorenzo2020rnn}\cite{kotseruba2020they}. PIE \cite{rasouli2019pie} utilized ConvLSTMs to integrate both spatial and temporal dynamics. Alternatively, other methods have leveraged 3D CNNs, such as 3D DenseNet \cite{saleh2019real}\cite{saleh2020spatio}, C3D \cite{tran2015learning}, and I3D \cite{carreira2017quo}, which inherently account for temporal reasoning. Additionally, some approaches have incorporated pose estimation as a feature, using 2D pose concatenation \cite{fang2018pedestrian}\cite{fang2019intention}\cite{wang2020estimating}, or 2D pose combined with Graph Convolutional Networks (GCNs) \cite{sucha2017pedestrian}\cite{cadena2019pedestrian}.
Adopting feature fusion techniques in recent works has improved prediction accuracy.  This process combines various features independently before integrating them. SF-GRU \cite{rasouli2020pedestrian} hierarchically fuses features at varying levels of complexity, while PCPA \cite{kotseruba2021benchmark} uses RNN encoders and temporal attention, followed by modality attention, to weigh the importance of each feature branch. This allows for the dynamic relevance of information features over time. To address the lack of global scene context in earlier methods, MaskPCPA \cite{yang2022predicting} and \cite{rasouli2022multi} integrate segmentation maps with attention-based fusion strategies for a more comprehensive understanding of the environment. Despite the growing focus on transformers, integrating multiple modalities\cite{zhao2021action}\cite{achaji2022attention}\cite{zhou2023pit}\cite{lorenzo2021capformer}\cite{zhang2023trep} for PIP remains challenging due to the need for large training datasets and complex architectures. While transformers show promise, RNN-based feature fusion methods often offer comparable performance with better interpretability, crucial for safety-critical applications.

\subsection{Pedestrian Trajectory prediction} 
%Pedestrian trajectory prediction has evolved significantly, transitioning from physics-based models to sophisticated data-driven approaches utilizing deep learning techniques. Early deep learning approaches in trajectory prediction were primarily deterministic, focusing on generating a single, most likely future path. Methods like PIETraj \cite{rasouli2019pie} employ an LSTM-based architecture that first estimates a pedestrian's crossing intention before refining trajectory predictions using past motion and ego-vehicle speed. Building on this, \cite{liu2022mdst} introduced spatiotemporal graph convolutional networks to capture complex interactions in the environment, while \cite{zhou2024pedestrian} combined trajectory features with knowledge-based pedestrian intentions to enhance prediction accuracy. MTNTraj \cite{yin2021multimodal} further advanced deterministic prediction by using a multimodal transformer network to fuse various data channels, including optical flow, at different granularities, capturing both broad and detailed motion dynamics.

%Pedestrian trajectory prediction has evolved from physics-based models to deep learning approaches. 
Early works in PTP employ vanilla RNN architectures like PIETraj \cite{rasouli2019pie}, which estimates crossing intentions which is then used to enhance trajectory predictions. More advanced techniques, such as spatiotemporal graph convolutional networks \cite{liu2022mdst}, dynamic-based deep models~\cite{wang2024pedestrian}, and knowledge-based intention models \cite{zhou2024pedestrian}, improved prediction accuracy. MTNTraj \cite{yin2021multimodal} further advanced deterministic prediction by using a multimodal transformer network to capture both high-level and fine-grained motion dynamics.

Recognizing the uncertainty in human motion, recent work has focused toward stochastic models that generate multiple plausible future trajectories. \cite{anderson2019stochastic} introduced a simulation method for diverse pedestrian trajectories, while BiTraP \cite{yao2021bitrap} uses a bi-directional decoder for long-term prediction accuracy and fully attention model presented in~\cite{rasouli2024novel}. SGNet \cite{wang2022stepwise}  incorporates goals at multiple temporal scales, further enhancing prediction. These models improve autonomous systems by offering a range of potential future scenarios, as highlighted in \cite{meng2024social}\cite{uhlemann2024evaluating}\cite{alghodhaifi2023holistic}. 
% \textcolor{blue}{[\textbf{is this connected to flow or disjoint component, if so we can remove]}Although trajectory prediction is well-studied in top-down views, ego-view research remains less explored.} (cite survey papers for ptp.)

\subsection{Datasets}

Early trajectory datasets like UCY~\cite{lerner2007crowds}, ETH~\cite{pellegrini2009eth}, and SSD~\cite{robicquet2016learning} capture pedestrian movement but use top-down views, making them unsuitable for studying vehicle-pedestrian interactions in ego-view. In contrast, ego-view datasets like nuScenes~\cite{caesar2020nuscenes}, Waymo~\cite{sun2020scalability}, and Argoverse2~\cite{wilson2021aroverse2} capture agent interactions but lack detailed pedestrian action annotations.
To enhance pedestrian action understanding, datasets like JAAD~\cite{rasouli2017they} and STIP~\cite{liu2020spatiotemporal} introduced binary intent annotations for predicting road-crossing behavior, with PSI~\cite{chen2021psi} adding dynamic intent annotations. However, these datasets often lack detailed ego-vehicle motion and explicit pedestrian action data. PIE~\cite{rasouli2019pie} and TITAN~\cite{malla2020titan} addressed these gaps by incorporating both, enabling the development of more context-aware models in structured settings.
%Recognizing the need for deeper pedestrian behavior analysis, datasets like JAAD~\cite{rasouli2017they} and STIP~\cite{liu2020spatiotemporal} introduced binary intent annotations to predict whether a pedestrian will cross the road. PSI~\cite{chen2021psi} further improved this with dynamic intent annotations. However, these datasets often lack detailed ego-vehicle motion data and explicit pedestrian action annotations. PIE~\cite{rasouli2019pie} and TITAN~\cite{malla2020titan} addressed these gaps by incorporating both, allowing researchers to build more context-aware models that account for pedestrian interactions with their environment in structured settings.

Despite advancements, existing datasets still lack in capturing complex and challenging scenarios, as illustrated in Fig.~\ref{fig:teaser}. Many lack lighting diversity—NightOwls~\cite{neumann2019nightowls} focuses only on nighttime, while PIE and TITAN cover only daytime. Additionally, datasets like Euro-PVI~\cite{bhattacharyya2021euro} and PePScenes~\cite{rasouli2020pepscenes} lack explicit pedestrian-vehicle interaction annotations, and there is limited data on occlusions and missing traffic signals or crosswalks in unstructured environments. %These gaps underscore the need for more comprehensive datasets.
%Despite advancements, existing datasets still face limitations in handling complex scenarios as shown in Fig.~\ref{fig:teaser}. Many lack diversity in lighting conditions—NightOwls~\cite{neumann2019nightowls} focuses on nighttime scenes, while PIE and TITAN cover only daytime settings. Additionally, datasets like PVI~\cite{bhattacharyya2021euro} and PePScenes~\cite{rasouli2020pepscenes} lack explicit annotations for pedestrian-vehicle interactions, and there is a scarcity of annotated data on occlusions and traffic signal or crosswalk absence in unstructured environments. These gaps highlight the need for more comprehensive datasets to address such challenges.
To address these gaps, we introduce a comprehensive dataset focused on pedestrians in unstructured scenarios
%, with annotations for bounding boxes, intentions, locations, ego-vehicle parameters, scene context, and interactions (see Table ~\ref{tab:dataset_comparison} for comparison with standard datasets). This holistic approach 
which enables the development of robust prediction models for navigating complex, real-world unstructured driving environments.

\section{The IDD-PeD Dataset}
In this section, we introduce our dataset and present details of the data collection, annotation and data statistics.

\begin{figure*}[!htbp]
    \centering
    \includegraphics[width=\linewidth]{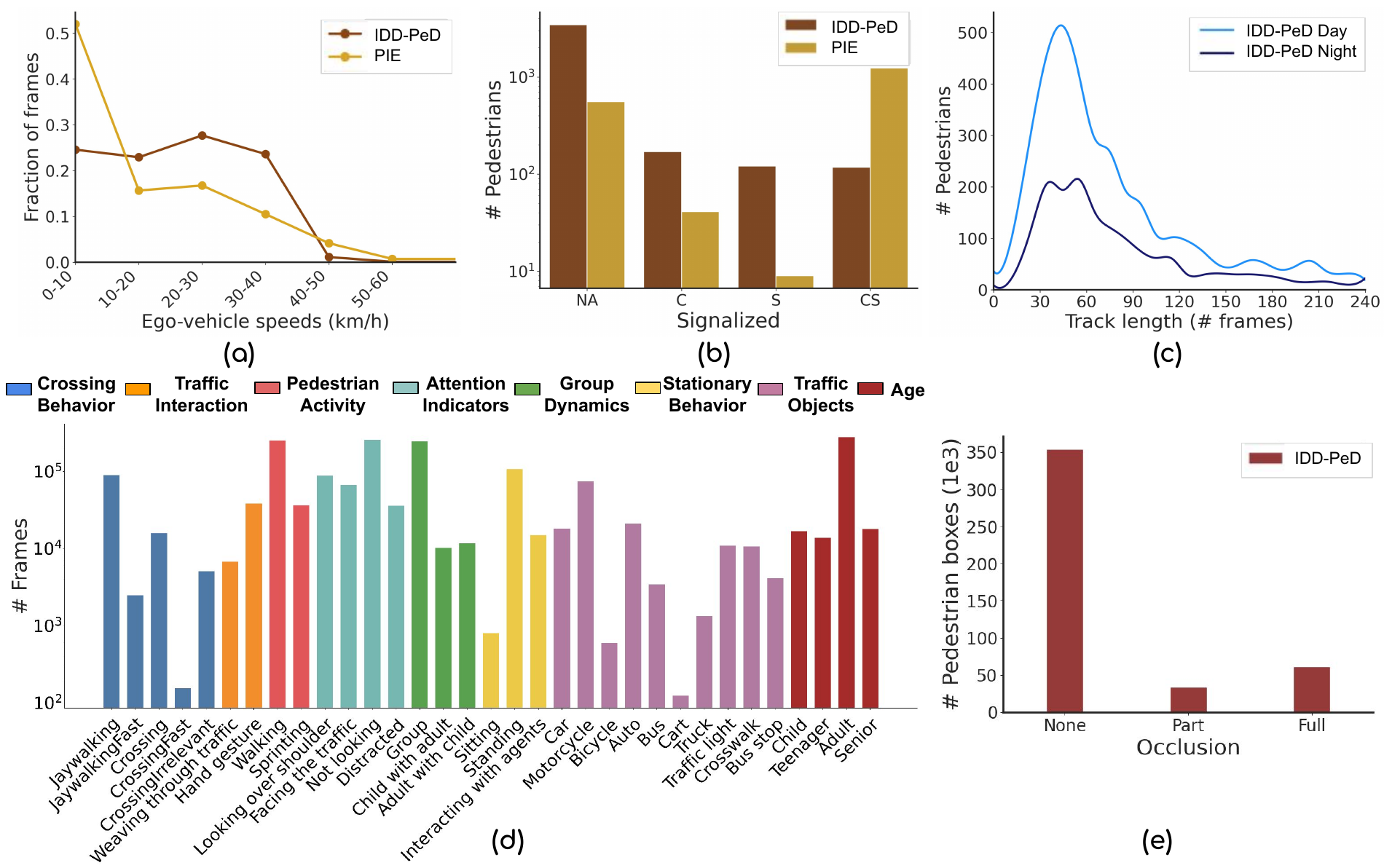}
    \caption{\textbf{Annotation instances and data statistics of IDD-PeD.} Distribution of (a) frame-level ego-vehicle speeds, (b) pedestrian at signalized types such as crosswalk (C), signal (S), crosswalk and signal (CS), and absence of crosswalk and signal (NA), (c)  pedestrians with track lengths at day and night, (d) frame-level different behavior analysis and traffic objects annotation, and (e) pedestrian occlusions.}
    \label{fig:datastats}
\vspace{-.4cm}
\end{figure*}

\subsection{Data Collection}
The dataset is curated from over $100$ hours of driving videos captured using either of the two ego-mounted cameras facing the front i.e., GoPro Hero $8$ Black at $30$ FPS and DDPAI X2SPro dashcam at $25$ FPS in HD format. Recordings from the GoPro camera are segmented into video clips of varying lengths, while dashcam recordings are consistently clipped into one-minute segments. We used an OBD sensor synchronized with the cameras which logs ego-motion data like acceleration, speed and GPS at $10$ Hz. 
The data collection was conducted in a densely populated urban area in South Asia, covering various weather conditions. 
%The scenes encompassed a wide diversity of pedestrian behaviours exhibited in chaotic and unstructured traffic conditions, including lighting conditions (day and night), pedestrian and ego-vehicle interactions, singalized and unsignalized traffic locations ranging from commercial roads with no pavements to residential areas with narrow streets and no crosswalk, partial, and no occlusion pedestrians in dense traffic scenes.
The scenarios captured a wide range of pedestrian behaviors in chaotic, unstructured traffic conditions. This includes both day and night lighting, pedestrian-vehicle interactions, and signalized and unsignalized traffic locations. The scenes vary from commercial roads without pavements to residential areas with narrow streets and no crosswalks, as well as dense traffic scenes with partial or full pedestrian occlusions.

\subsection{Data Annotations}
 Our dataset provides five types of annotations for pedestrians requiring the ego-vehicle's attention. We used the CVAT tool\footnote{https://app.cvat.ai/} for manually annotating bounding boxes and other annotations, while 2D pose annotations were automatically extracted using  MMPose~\cite{mmpose2020}.
 % The annotators manually annotated all the frames.
 We provide examples of the five types of annotations in the supplementary video. 

%\noindent \textbf{Spatial annotations:} Bounding boxes are provided for detecting and tracking pedestrians, as well as relevant traffic objects such as vehicles, signals, bus stations, and crosswalks, which influence pedestrian movement~\cite{vasudevan2020pedestrian,zhang2023analysis}. The IDD-PeD dataset includes $205,145$ annotated frames with $494,854$ pedestrian bounding boxes and $190,395$ bounding boxes for other traffic objects like vehicles, traffic signals, bus-stations, and crosswalks. Due to occlusions in our unstructured and dense data (Fig.~\ref{fig:teaser}), all bounding boxes include occlusion information.
\noindent \textbf{Spatial annotations:} The IDD-PeD dataset provides bounding boxes for pedestrians and relevant traffic objects like vehicles, traffic signals, bus stations, and crosswalks, which influence pedestrian movement~\cite{vasudevan2020pedestrian,zhang2023analysis}. It includes $205,145$ annotated frames, with $494,854$ pedestrian boxes and $190,395$ boxes for other traffic objects. All bounding boxes contain occlusion information (see Fig.~\ref{fig:datastats}(e)).

%\noindent \textbf{Behavioral annotations:} are frame-level annotation of attributes that capture each pedestrian behavior. In unstructured environments, pedestrians display a wide variety of behaviors that can be analyzed to understand their intentions, predict future actions, and serve as cues for forecasting their future movements~\cite{kotseruba2020they,yao2021coupling}. 
\noindent \textbf{Behavioral annotations:} are frame-level annotations that capture pedestrian behavior in unstructured environments, providing cues for understanding intentions, predicting actions, and forecasting movements~\cite{kotseruba2020they,yao2021coupling}.
In our dataset, we categorize behavioral attributes into six distinct classes as shown in Fig.~\ref{fig:datastats}(d) comprising of 
%(i) \textit{Crossing behavior} captures pedestrian actions while moving across the road. For example, "Crossing" refers to a pedestrian following traffic rules, such as using a crosswalk when the pedestrian signal is green. In contrast, "Jaywalking" occurs when a pedestrian crosses the road illegally, such as when the vehicle signal is green. This behavior also reflects the level of risk pedestrians face while crossing. 
(i) \textit{Crossing Behavior} referring to pedestrian actions when crossing the road. ``Crossing" follows traffic rules, like using a crosswalk with a green signal, while ``Jaywalking" occurs when crossing illegally~\cite{Mukoya2024Jaywalker}, such as against a vehicle signal, indicating varying risk levels.
%
% (ii) \textit{Traffic engagement} refers to how pedestrians interact with vehicular traffic. In unstructured environments, this often includes non-verbal communication with road users, such as using a "hand gesture" to signal intentions, or "weaving through traffic", where pedestrians navigate intricate paths between slow-moving vehicles. 
(ii) \textit{Traffic Interaction} show pedestrian interactions with vehicles~\cite{muktadir2024adaptive}, including non-verbal cues like ``hand gestures" to signal intentions or ``weaving through traffic", where pedestrians navigate between slow-moving vehicles.
(iii) \textit{Pedestrian Activity} records pedestrian behaviors when they are moving along the side of the road like ``walking", or ``sprinting". 
%
% (iv) \textit{Attention indicators} reflect behaviors related to a pedestrian’s focus in relation to the ego-vehicle’s direction. For example, a pedestrian "looking over shoulder" indicates awareness of the approaching vehicle, while a "distracted" pedestrian, such as one looking at their phone, shows a lack of attention to the vehicle’s presence. 
(iv) \textit{Attention Indicators} capture a pedestrian's focus relative to the ego-vehicle. For instance, ``looking over shoulder" signals awareness of the vehicle, while being ``distracted" (e.g., using a phone) shows inattention.
%
% (v) \textit{Group dynamics} captures the behaviors of pedestrians moving in groups, a common occurrence in dense environments. For example, a "child with an adult" often exhibits different movement patterns than individuals walking alone, and a "crowd" of pedestrians may move as a cohesive unit, affecting their interactions with traffic. 
(v) \textit{Group Dynamics} shows pedestrian behaviors in groups, common in dense areas. For example, a ``child with an adult" shows different movement patterns than someone walking alone, while a ``crowd" may move as a unit, influencing traffic interactions.
%
% (vi) \textit{Stationary behavior} refers to the actions of pedestrians who remain in one place, often interacting with their surroundings. For example, in a bustling street environment, pedestrians frequently engage with hawkers on the side of the road, demonstrating this stationary behavior through their "Communicating with agents" interactions.
(vi) \textit{Stationary Behavior} describes pedestrians staying in one place, often interacting with their surroundings. In busy unstructured streets, this includes engaging with hawkers or others, exemplified by ``Communicating with agents" interactions, common in dense urban environments.
%These behavioral attributes help understand individual pedestrian actions as well as their interplay with the surroundings. Each pedestrian is annotated with the appropriate attribute from each of these classes, resulting in six behavioral attributes per frame per pedestrian. 
% put group size in what?

\noindent \textbf{Scene annotations:} describe the environmental contextual information around the pedestrian, such as the road type (main road, streets), intersection type (mid-block, turns), signalized type (crosswalk, traffic signal), and time of day (day, night). Scene annotations are crucial in unstructured environments as they provide context for understanding chaotic pedestrian movements. By associating behaviors with environmental factors, these annotations help uncover patterns behind random pedestrian actions. For example, in a busy street with pedestrians weaving through traffic, scene annotations can link a pedestrian’s sudden change in direction to a nearby bus-stop or a crosswalk, revealing motives behind their movement.
%Scene annotations are essential in unstructured environments, where they provide critical context to interpret seemingly chaotic pedestrian movements. By linking behaviors to specific environmental factors, these annotations help reveal patterns and motivations behind pedestrian actions that may otherwise appear random or unpredictable.  \\
% \textcolor{blue}{any other elements like weather, traffic signals, time of day, etc be specific} 
% For relevant traffic objects, namely vehicles, traffic signals, bus-stations and crosswalks we provide annotations in the form of bounding boxes and attributes. These were chosen based on the influence they have on pedestrian actions. For instance, pedestrians appear to perceive a higher risk when facing larger vehicles, as evidenced by their preference for larger gaps in the traffic flow as the vehicle size increased~\cite{vasudevan2020pedestrian}

\noindent \textbf{Interaction annotations:} capture how pedestrians and vehicles interact in rule-flexible unstructured environments, where both may alter their speed or direction. Each pedestrian track is labeled with a binary indicator showing whether an interaction with the ego-vehicle occurred and a label indicating the nature of the interaction. For example, in a congested intersection, if a pedestrian abruptly changes their path to avoid an approaching vehicle, this interaction is annotated to highlight the challenges in predicting and responding to such dynamic scenarios.

%\noindent \textbf{Interaction annotations:} With rule-flexible traffic situations in unstructured environments often contain vehicle-pedestrian interactions, where either or both change their motion (i.e., speed or direction). We annotate every pedestrian track with a binary label indicating the presence or absence of an interaction with the ego-vehicle and show \textcolor{blue}{"show" --> how is this done not clear} empirically the challenges associated with such interactions. \\

\noindent \textbf{Location annotations:} provide spatial context for pedestrian movements in unstructured environments. Attributes such as ``near divider", ``side of the road", and ``near crosswalk" help situate pedestrians within their surroundings. When paired with behavior attributes like ``looking over shoulder" or ``distracted", these location indicators enhance predictions of pedestrian intent. For example, a pedestrian ``near crosswalk" who is ``distracted" might be at higher risk of crossing unpredictably, improving safety analysis and prediction models.

\subsection{Data Statistics}
Fig.~\ref{fig:datastats} gives a detailed overview of the data statistics. Our dataset contains over $5,000$ annotated pedestrians distributed across continuous videos with a varying length of $45$ seconds to $10$ minutes.
Our dataset features an even distribution of ego-vehicle speeds, with a median of $30$ km/h (see Fig.~\ref{fig:datastats}(a)). In contrast, the PIE dataset includes many frames where the ego-vehicle is stationary, simplifying pedestrian behavior modeling due to the static camera. Pedestrian encounters in our dataset are more frequent in areas without signals or crosswalks (NA), while in structured environments like PIE, presence of crosswalks and traffic signals (CS) leads to smoother pedestrian movement (see Fig.~\ref{fig:datastats}(b)). The pedestrian tracks have a median length of approximately $60$ frames and cover both day and night scenarios (see Fig.~\ref{fig:datastats}(c)). We use the PIE~\cite{rasouli2019pie} occlusion labeling convention 
%: None for \textless25\% occlusion, Part for 25-75\% occlusion, and Full for \textgreater75\% occlusion of the bounding box area 
(see Fig.~\ref{fig:datastats}(e)). The dataset is split into training ($70\%$) and testing ($30\%$) sets.

\section{Experiments}
% Details about what experiments have been conducted and for what purpose.
In this section, we discuss the experimental settings, evaluation of the proposed IDD-PeD dataset on pedestrian intention and trajectory prediction, and further analysis. 
%We show experimental results for the following two tasks to demonstrate the performance of state-of-the-art methods and their limitations given the nature of scenarios commonly encountered in highly unstructured environments such as ours: \\

\subsection{Experimental Settings}
\noindent \textbf{Baseline Methods.} \noindent We evaluate eight PIP baseline methods on our dataset, namely, Static \cite{simonyan2014very}, SingleRNN \cite{kotseruba2020they}, I3D \cite{carreira2017quo}, C3D \cite{tran2015learning}, ConvLSTM \cite{shi2015convolutional}, SF-GRU \cite{rasouli2020pedestrian}, PCPA \cite{kotseruba2021benchmark}, and MaskPCPA \cite{yang2022predicting}. In PTP, we evaluate two deterministic baselines-- PIETraj \cite{rasouli2019pie}, MTN \cite{yin2021multimodal} and two stochastic baselines-- BiTraP \cite{yao2021bitrap}, SGNet \cite{wang2022stepwise}. These baselines are then evaluated across the key challenges including \textit{occlusions}, \textit{illumination changes}, \textit{unsignalized} scenes, and vehicle-pedestrian \textit{interactions}.
  %The baseline methods used collectively cover a broad spectrum of deep learning approaches, from basic spatial analysis to complex, multi-modal integration and including and excluding attention mechanisms. One or more visual and auxiliary features such as images, bounding boxes, poses, ego-vehicle speeds etc. are used by the different baseline methods. 

\begin{table}%[h]
\centering
\caption{Evaluation of \textbf{PIP baselines} on JAAD, PIE and our datasets.}
\label{tab:model_performance_across_datasets}
\footnotesize 
\setlength{\tabcolsep}{5pt} 
\renewcommand{\arraystretch}{1.2}
\begin{tabular}{r |cc | cc | cc}
\hline
\multirow{2}{*}{Model} & \multicolumn{2}{c|}{JAAD~\cite{rasouli2017they}} & \multicolumn{2}{c|}{PIE~\cite{rasouli2019pie}} & \multicolumn{2}{c}{IDD-PeD (ours)} \\
\cline{2-7}
& \multicolumn{1}{c}{AUC} & \multicolumn{1}{c|}{$F_1$} & \multicolumn{1}{c}{AUC} & \multicolumn{1}{c|}{$F_1$} & \multicolumn{1}{c}{AUC} & \multicolumn{1}{c}{$F_1$} \\
\hline
Static~\cite{simonyan2014very} & 0.75 & 0.55 & 0.60 & 0.41 & 0.56 & 0.21 \\
ConvLSTM~\cite{shi2015convolutional} & 0.57 & 0.32 & 0.55 & 0.39 & 0.50 & 0.15 \\
C3D~\cite{tran2015learning} & 0.81 & 0.65 & 0.67 & 0.52 & 0.59 & 0.25 \\
I3D~\cite{carreira2017quo} & 0.74 & 0.63 & 0.73 & 0.62 & 0.61 & 0.30 \\
SF-GRU~\cite{rasouli2020pedestrian} & 0.77 & 0.53 & 0.79 & 0.69 & 0.59 & 0.26 \\
SingleRNN~\cite{kotseruba2020they} & 0.77 & 0.54 & 0.77 & 0.67 & 0.61 & 0.31 \\
\rowcolor[gray]{0.9}PCPA~\cite{kotseruba2021benchmark} & 0.79 & 0.56 & 0.86 & 0.77 & $\mathbf{0.71}$ & $\mathbf{0.33}$ \\
MaskPCPA~\cite{yang2022predicting} & 0.82 & 0.63 & 0.86 & 0.80 & 0.61 & 0.25 \\
\hline
\end{tabular}
\label{tab:pip_models}
\vspace{-.4cm}
\end{table}

\noindent \textbf{Implementation Details.} We implement baseline methods on Intel Xeon E$5$-$2640$ v$4$ processor with multiple Nvidia RTX $2080$ Ti GPUs. 
%The training and testing split of our dataset is $70\%$ and $30\%$.
We fix an observation period of $0.5$s and a time-to-event varying between $1-2$s for PIP while in PTP we predict future paths up to $1.5$s.
For a fair comparison, we adopted the same observation and time settings as in \cite{kotseruba2021benchmark} for PIP and \cite{rasouli2019pie} for PTP. We use RAFT~\cite{teed2020raft} to extract optical flow information, used by the MTN \cite{yin2021multimodal} method. We ensured that for all the crossing cases, both the observation period and the time-to-event precede the pedestrian's road crossing initiation.
% \textcolor{blue}{\textbf{[May be this is not essential or can be reduced into single sentence]}, train-test split and evaluation metrics. For a fair comparison, we have followed the same observation and time-to-event settings for our dataset and have chosen a train test split of \_/\_. The time-to-event denotes the buffer time post the end of the observation period so that the vehicle has enough time~\cite{schmidt2009pedestrians} to plan it's maneuver accordingly. In all cases of crossing, the observation period $t_o$ and the time-to-event are both before the pedestrian even starts to cross the road.} 
For methods that use ego-vehicle speed, we report results on PIE and our dataset, since JAAD doesn't provide explicit ego-vehicle speeds. 
%while excluded in JAAD since JAAD doesn't provide ego-vehicle speeds. 
%For PTP, we consider an observation period of $0.5$s and predict future trajectories up to $1.5$s, adopting . 

% The predicted trajectories are evaluated in two forms: bounding box coordinates and centers, using the commonly used evaluation metrics in the literature \cite{rasouli2019pie}\cite{yin2021multimodal}\cite{yao2021bitrap}\cite{wang2022stepwise}, namely Average Displacement Error (MSE), Center MSE (C-MSE), Center Final Displacement Error (C-FDE), all measured in squared pixels. Bounding box MSE represents the mean square error between predicted and ground-truth trajectories averaged across the prediction period. C-FDE represents the mean square error between the centers of the predicted and ground-truth bounding boxes at the final prediction timestep.

\noindent \textbf{Evaluation Metrics.} We evaluate PIP using AUC and $F_1$ scores, and PTP using Mean Square Error (MSE), Center MSE (C-MSE), Center Final Mean Square Error (CF-MSE), all measured in squared pixels. 
For stochastic methods, we report the best-of-20 results, consistent with~\cite{gupta2018social}.

\subsection{Intention Prediction Results}
%\noindent We've used the following baseline methods to benchmark on our dataset - Static \cite{simonyan2014very}, SingleRNN \cite{kotseruba2020they}, I3D \cite{carreira2017quo}, C3D \cite{tran2015learning}, ConvLSTM \cite{shi2015convolutional}, SF-GRU \cite{rasouli2020pedestrian}, PCPA \cite{kotseruba2021benchmark}, MaskPCPA \cite{yang2022predicting}. The baseline methods used collectively cover a broad spectrum of deep learning approaches, from basic spatial analysis to complex, multi-modal integration and including and excluding attention mechanisms. One or more visual and auxiliary features such as images, bounding boxes, poses, ego-vehicle speeds etc. are used by the different baseline methods. 
In Table~\ref{tab:pip_models}, we report experiments of the PIP baselines on PIE, JAAD and our datasets. We observe that existing models like SF-GRU, SingleRNN, and PCPA are less robust to unstructured environmental changes specifically in the context of our dataset. 
%\textcolor{blue}{\textbf{[Need to write better the imbalance issue of our dataset vs other dataset]} Given the imbalance between crossing and not crossing samples in our dataset, with majority of samples being not crossing, unlike in PIE and JAAD the data is balanced, leads to improvement in accuracy.} To overcome this imbalance in real-world applications, 
%Keeping in mind the concern of false negatives for real-world pedestrian safety applications, we focus on $AUC$ and $F_1$ metrics. 
PCPA shows a superior performance among the baselines with an AUC performance drop of $15\%$ and $8\%$, and $F_1$ of $44\%$ and $23\%$, between our dataset, and PIE and JAAD respectively. Surprisingly, the recent MaskPCPA method, which incorporates global context via segmentation maps while achieving SOTA on PIE and JAAD datasets significantly underperforms on our dataset. One possible reason is the parameter-heavy architecture of PCPA, which has $31$M parameters compared to $3$M in MaskPCPA. This larger capacity allows PCPA to better model the noisy and complex patterns in unstructured settings.

\subsection{Trajectory Prediction Results}

\begin{table}[t]
\centering
\caption{Evaluation of \textbf{PTP baselines} on JAAD, PIE and our datasets. We report MSE results at $1.5$s. ``-" indicates no results as PIETraj needs explicit ego-vehicle speeds. See supplementary video for MSE with $0.5$s and $1$s prediction periods.
%are the MSEs over the center of the bounding boxes for the entire predicted sequence and only the last time step respectively.
}
\setlength{\tabcolsep}{1pt}
\begin{tabular}{@{}c@{}r| ccc|ccc|ccc@{}}
\hline
\multicolumn{2}{@{}l|}{\multirow{3}{*}{{Method}}} & \multicolumn{3}{c|}{PIE~\cite{rasouli2019pie}} & \multicolumn{3}{c|}{JAAD~\cite{rasouli2017they}} & \multicolumn{3}{c}{IDD-PeD (ours)} \\
\cline{3-11}
\multicolumn{2}{@{}l|}{} & \scriptsize{MSE} & \scriptsize{C-MSE} & \scriptsize{CF-MSE} & 
   \scriptsize{MSE} & \scriptsize{C-MSE} & \scriptsize{CF-MSE} &
  \scriptsize{MSE} & \scriptsize{C-MSE} & \scriptsize{CF-MSE} \\
\hline
 & PIETraj~\cite{rasouli2019pie} & 559 & 520 & 2162 & - & - & - & 2181 & 1979 & 8960 \\
 
\rowcolor[gray]{0.9} & MTN~\cite{yin2021multimodal} & $\mathbf{444}$ & $\mathbf{414}$ & $\mathbf{1627}$ & $\mathbf{1005}$ & $\mathbf{951}$ & $\mathbf{4010}$ & $\mathbf{1652}$ & $\mathbf{1489}$ & $\mathbf{6668}$ \\
\hdashline
 & BiTraP~\cite{yao2021bitrap} & 102 & 81 & 261 & 222 & 177 & 565 & 338 & 200 & 638 \\
\rowcolor[gray]{0.9} & SGNet~\cite{wang2022stepwise} &  $\mathbf{88}$ & $\mathbf{66}$ & $\mathbf{206}$ & $\mathbf{197}$ & $\mathbf{146}$ & $\mathbf{443}$ & $\mathbf{310}$ & $\mathbf{190}$ & $\mathbf{531}$ \\
\hline
\end{tabular}
\label{tab:ptp_models}
\vspace{-.4cm}
\end{table}

% \textcolor{blue}{\textbf{[Not appropriate to start with Further analysis table, start directly with Table III. Also I do not think MSE is calculated in percentage writing 31\%, 164\%, etc do they have any meaning, find better way to distinguish when it comes to MSE]}} 

In Table~\ref{tab:ptp_models}, we report the MSE results of PTP baselines at various prediction intervals on PIE, JAAD and our datasets. We demonstrate that existing methods such as PIETraj, MTN that are deterministic, BiTraP and SGNet that are stochastic, perform with increased error rates when evaluated on our dataset which indicates a lack of robustness to diverse pedestrian behaviours in unstructured environments. Despite not achieving the best results on our dataset, the stochastic SGNet model has the best overall results, with a performance drop of $222$ and $113$ MSE between ours and the PIE and JAAD datasets respectively. Similarly, the deterministic MTN has the best results, with a performance drop of $1,208$ and $647$ MSE between ours, and PIE and JAAD respectively.
The more substantial increase in error rates for PIETraj and MTN on our dataset might be attributed to their heavy reliance on visual features, that could be more susceptible to noise and variability whereas models like SGNet, that use a bounding-box based approach may offer some resilience to the visual complexities of unstructured environments.

\begin{table}[t]
\centering
\caption{Evaluation of PIP baselines on \textbf{Occlusions, Signalized Types, Illumination Changes}, and \textbf{Vehicle-Pedestrian Interactions} in our dataset. We report AUC scores. 
%For this comparison, to account for the imbalance of positive and negative samples, we have sampled negative instances to match the number of positive samples using a deterministic seed.
}
\label{tab:comparison}
\setlength{\tabcolsep}{3pt}
\begin{tabular}{@{}r|cc|cc|cc|cc@{}}
\hline
\multirow{2}{*}{\textbf{Method}} & \multicolumn{2}{c|}{\textbf{Occlusion}} & \multicolumn{2}{c|}{\textbf{Signalized}} & \multicolumn{2}{c|}{\textbf{Illumination}} & \multicolumn{2}{c@{}}{\textbf{Interaction}} \\
 & w/o & w & Yes & No & Day & Night & w/o & w \\
\hline
% Static & - & - & 0.45 & 0.59 & 0.65 & 0.42 & 0.57 & 0.63 \\
SingleRNN~\cite{kotseruba2020they} & 0.63 & 0.61 & 0.70 & 0.62 & 0.63 & 0.58 & 0.65 & 0.58 \\
% ConvLSTM & - & - & 0.54 & 0.49 & 0.58 & 0.32 & 0.52 & 0.52 \\
% C3D & - & - & 0.42 & 0.60 & 0.61 & 0.45 & 0.58 & 0.58 \\
I3D~\cite{carreira2017quo} & 0.60 & 0.57 & 0.48 & 0.64 & 0.65 & 0.51 & 0.64 & 0.60 \\
SF-GRU~\cite{rasouli2020pedestrian} & 0.60 & 0.59 & 0.60 & 0.59 & 0.62 & 0.53 & 0.61 & 0.59 \\
\rowcolor[gray]{0.9}PCPA~\cite{kotseruba2021benchmark} & $\mathbf{0.72}$ & $\mathbf{0.71}$ & $\mathbf{0.66}$ & $\mathbf{0.72}$ & $\mathbf{0.74}$ & $\mathbf{0.65}$ & $\mathbf{0.73}$ & $\mathbf{0.69}$ \\
% MaskPCPA & - & - & 0.65 & 0.60 & 0.63 & 0.56 & 0.54 & 0.69 \\
\hline
\end{tabular}
\label{tab:pip_comparisons}
\vspace{-.4cm}
\end{table}

\subsection{Further Analysis of Our Dataset Challenges}
\label{sec:further_analysis}
% \textcolor{blue}{For PIP and PTP 4 key factors tables and explanations need to go here.}
Here, we examine the diversity and richness of our dataset, with quantitative and qualitative results for PIP and PTP.

% \textbf{Occlusions}

% \textbf{Illumination changes}

% \textbf{Unsignalized locations}

% \textbf{Interactions}

\noindent \textbf{Intention Prediction.} In Table~\ref{tab:pip_comparisons}, we report AUC of the best performing models across key challenges present in our dataset. We observe that the performance of all the baselines drops when a part of the pedestrian sequence is \textit{occluded}. While occlusion is important, its impact diminishes when there are enough non-occluded frames in the sequence.
We also observe that the \textit{nightlight} performance drops by up to $14\%$ over \textit{daylight} scenes on all baselines making our dataset more challenging. This is attributed to low and in-homogeneous illumination, which hinders the baselines that heavily rely on visual features. Similarly, the presence of \textit{vehicle-pedestrian interactions} significantly impacted the performance of the baselines causing a performance drop of up to $4\%$ compared to situations with no interactions. This can be attributed to a sudden change in the decision of the pedestrian to cross or not cross which may have caused the interaction. 
%%%%%% Commented out by Ruthvik
% Interestingly, experiments on \textit{signalized type} gives PCPA to be superior in unsignalized locations, with a performance gain of $6\%$ over signalized locations. Unstructured settings are commonly accompanied by a non-adherence to traffic rules (see Fig.~\ref{fig:Ours_PIE}), high instances of jaywalkers (see Table~\ref{tab:dataset_comparison}), and therefore environmental cues like crosswalks or signals may not always correlate with pedestrian behaviours.
%%%%%% Commented out by Ruthvik

\noindent \textbf{Trajectory Prediction.} In Table~\ref{tab:ptp_comparisons}, we demonstrate the MSE values at $1.5$s across baseline methods, highlighting the challenges within our dataset. The presence of \textit{occlusion} results in a significant MSE increase of $603$ and $349$ for PIETraj and MTN, respectively, while BiTraP and SGNet show a rise of $133$ and $122$ MSE, respectively.
The performance across \textit{illumination} scenarios is also evident for PIETraj and MTN, which exhibit a performance drop during the nightlight over the daylight. Interestingly, BiTraP and SGNet show less sensitivity to this challenge due to their reliance on bounding boxes rather than raw images. %information rather than raw images, making them less susceptible to visual changes in low-light conditions. 
The largest performance drop comes from scenarios involving \textit{pedestrian-vehicle interactions} with a drop by up to $354$ and $1,340$ MSE for SGNet and MTN. Interactions often involve instantaneous movements by pedestrians and modelling this is critical in unstructured scenarios. Finally, the presence of \textit{signalized} infrastructure correlates with improved model performance in trajectory prediction that can be attributed to fewer interactions and smoother pedestrian flow. 

\begin{figure}[h]
    \centering
    \includegraphics[width=\columnwidth]{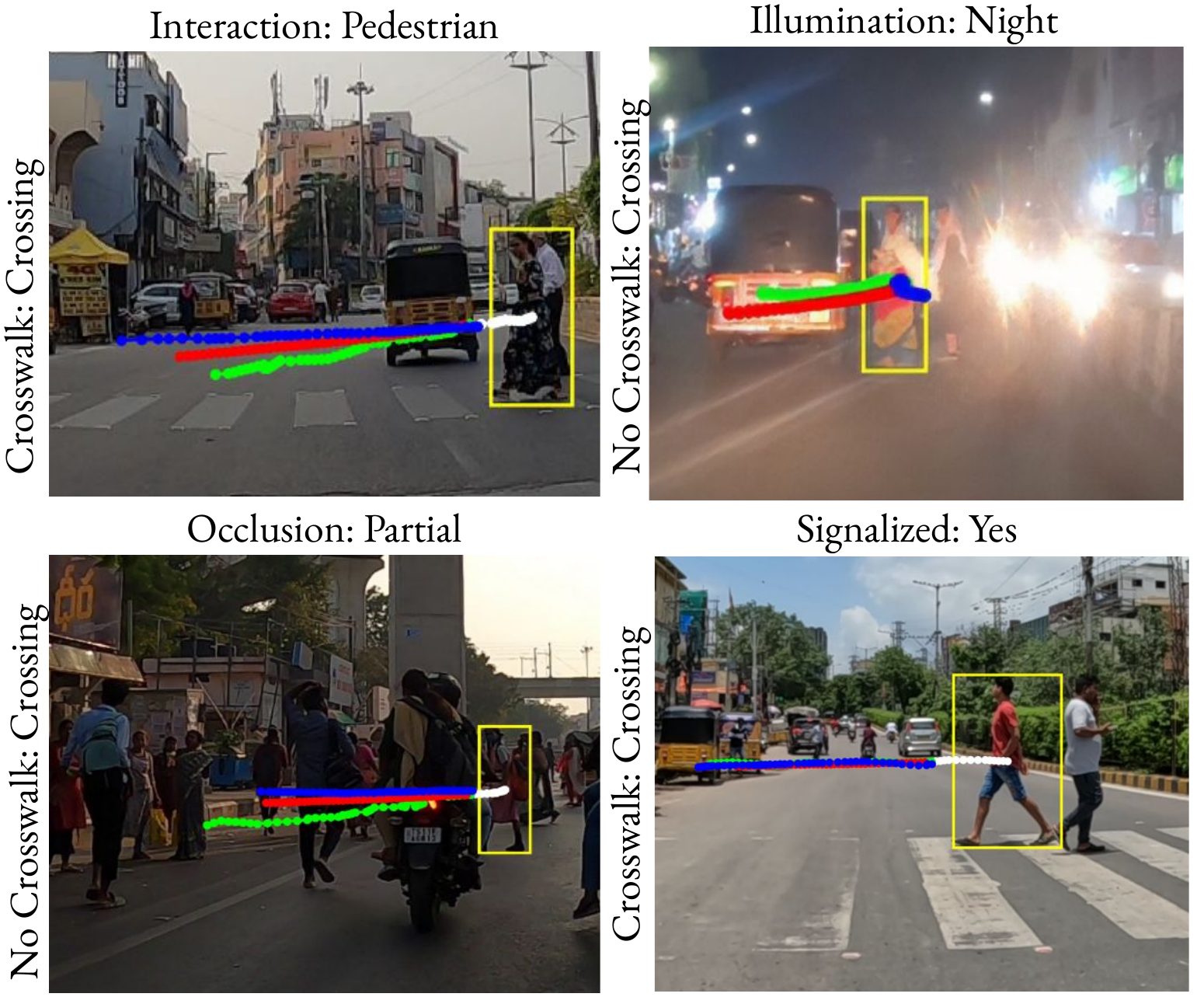}
    \caption{\textbf{Qualitative evaluation of the best and worst PTP models on our dataset.} \textcolor{red}{Red: SGNet}, \textcolor{blue}{Blue: PIETraj}, \textcolor{green}{Green: Ground truth}, White: Observation period. To better illustrate and highlight key factors in PIP and PTP methods, a qualitative analysis will be provided in the supplementary video.}
    \label{fig:ptp_qualitative}
    \vspace{-.3cm}
\end{figure}

\noindent \textbf{Qualitative Analysis.} We illustrate qualitative evaluation of the best (SGNet) and the worst (PIETraj) PTP models on our dataset in Fig.~\ref{fig:ptp_qualitative}. 
\textit{Interaction:} As the ego-vehicle approaches the crossing pedestrian, the pedestrian exhibits movement away from the ego-vehicle, a behavior both SGNet and PIETraj fail to capture, likely due to the presence of a crosswalk. 
\textit{Nightlight:} PIETraj does not predict the pedestrian’s crossing, possibly due to low visibility and uneven lighting. \textit{Partial occlusion:} Both models struggle to predict the pedestrian's future location, likely because of occlusion during the observation period. \textit{Signalized type:} Both models perform well in signalized settings, where smooth motion and the ego-vehicle at rest simplify prediction.
% \textit{Interaction:} With pedestrian crossing and moving away from the ego-vehicle in motion towards the pedestrian, both SGNet and PIETraj fail to account for the away movement, possibly due to the presence of a crosswalk. \textit{Nightlight:} PIETraj does not predict the crossing motion of the pedestrian, which can be possibly be attributed to low visibility and non-homogeneous illumination. \textit{Partial occlusion:} Both models are unable to anticipate the future location of the pedestrian, possibly due to the presence of occlusion in the observation period. \textit{Signalized type:} Both models perform well owing to smooth motion in the signalized location with the ego-vehicle at rest.  %This is a prevalent feature in structured environments, which makes modelling pedestrian behavior easier.

\begin{table}[t]
\centering
\caption{Evaluation of PTP baselines on
 \textbf{Occlusions, Signalized Types, Illumination Changes}, and \textbf{Vehicle-Pedestrian Interactions} in our dataset. We reported MSE at $1.5$s. 
 %We report the best-of-20 results for the stochastic methods.
 }
% \caption{PTP Performance comparison of different factors on our dataset}
\label{tab:comparison}
\setlength{\tabcolsep}{4pt} 
\begin{tabular}{l|cc|cc|cc|cc}
\hline
\multirow{2}{*}{\textbf{Method}} & \multicolumn{2}{c|}{\textbf{Occlusion}} & \multicolumn{2}{c|}{\textbf{Signalized}} & \multicolumn{2}{c|}{\textbf{Illumination}} & \multicolumn{2}{c}{\textbf{Interaction}} \\
 & w/o & w & Yes & No & Day & Night & w/o & w \\
\hline
PIETraj~\cite{rasouli2019pie} & 2055 & 2658 & 1434 & 2374 & 2190 & 2294 & 1561 & 4127 \\
MTN~\cite{yin2021multimodal} & 1565 & 1914 & 1215 & 1765 & 1633 & 1704 & 1328 & 2668 \\
\hdashline
BiTraP~\cite{yao2021bitrap} & 301 & 434 & 296 & 341 & 339 & 310 & 249 & 610 \\
\rowcolor[gray]{0.9}SGNet~\cite{wang2022stepwise} & $\mathbf{289}$ & $\mathbf{411}$ & $\mathbf{281}$ & $\mathbf{329}$ & $\mathbf{330}$ & $\mathbf{301}$ & $\mathbf{227}$ & $\mathbf{581}$ \\
\hline
\end{tabular}
\label{tab:ptp_comparisons}
\vspace{-.4cm}
\end{table}

% \begin{table}[t]
% \centering
% \caption{PIP Performance comparison of different factors on our dataset}
% \label{tab:comparison}
% \setlength{\tabcolsep}{4pt}
% \begin{tabular}{@{}l|cc|cc|cc|cc@{}}
% \hline
% \multirow{2}{*}{Method} & \multicolumn{2}{c|}{\textbf{Occlusion}} & \multicolumn{2}{c|}{\textbf{Signalized}} & \multicolumn{2}{c|}{\textbf{Illumination}} & \multicolumn{2}{c@{}}{\textbf{Interaction}} \\
%  & w/o & w & Yes & No & Day & Night & w/o & w \\
% \hline
% Static & - & - & 0.45 & 0.59 & 0.65 & 0.42 & 0.57 & 0.63 \\
% SingleRNN & - & - & 0.70 & 0.62 & 0.63 & 0.58 & 0.65 & 0.58 \\
% ConvLSTM & - & - & 0.54 & 0.49 & 0.58 & 0.32 & 0.52 & 0.52 \\
% C3D & - & - & 0.42 & 0.60 & 0.61 & 0.45 & 0.58 & 0.58 \\
% I3D & - & - & 0.48 & 0.64 & 0.65 & 0.51 & 0.64 & 0.60 \\
% SF-GRU & - & - & 0.60 & 0.59 & 0.62 & 0.53 & 0.61 & 0.59 \\
% PCPA & - & - & 0.66 & 0.72 & \textbf{0.74} & \textbf{0.65} & 0.73 & 0.69 \\
% MaskPCPA & - & - & 0.65 & 0.60 & 0.63 & 0.56 & 0.54 & 0.69 \\
% \hline
% \end{tabular}
% \label{tab:pip_models_comparison}
% \end{table}

\section{Conclusions}

In this work, we introduced IDD-PeD, a large pedestrian dataset with detailed behavioral annotations designed for unstructured traffic challenges, including occlusions, signalized types, vehicle-pedestrian interactions, and illumination variations.
Extensive experiments on state-of-the-art pedestrian intention and trajectory prediction baseline methods on our dataset higlights its complexity, with both quantitative and qualitative evaluations revealing significant performance degradation compared to existing datasets. 

This dataset provides key insights into vehicle-pedestrian interactions, aiding autonomous systems in anticipating pedestrian behavior. Combined with technological advancements, it aims to enhance navigation and ensure safety in unstructured and unpredictable traffic environments.

{\small
\bibliographystyle{ieeetr}
\bibliography{egbib}
}
\end{document}